\documentclass[letterpaper, 10 pt, journal, twoside]{IEEEtran}
\usepackage{amsmath,amsfonts}
\usepackage{algorithmic}
\usepackage{algorithm}
\usepackage{array}
\usepackage[caption=false,font=normalsize,labelfont=sf,textfont=sf]{subfig}
\usepackage{textcomp}
\usepackage{stfloats}
\usepackage{url}
\usepackage{verbatim}
\usepackage{graphicx}
\usepackage{cite}
\usepackage{subcaption}
\usepackage{booktabs}
\usepackage{multirow}
\usepackage{xcolor}
\usepackage{adjustbox}
\usepackage{enumitem}

\usepackage{hyperref}
\usepackage{cleveref}
\crefname{figure}{Figure}{Figures}
\crefname{table}{Table}{Tables}
\crefname{section}{Section}{SectionS}

\begin{document}

\markboth{IEEE Robotics and Automation Letters. Preprint Version. Accepted January, 2025}
{Leng \MakeLowercase{\textit{et al.}}: BEVCon: Advancing Bird's Eye View Perception with Contrastive Learning} 

\author{
Ziyang Leng$^{1}$, Jiawei Yang$^{2\dagger}$, Zhicheng Ren$^{3\dagger}$, and Bolei Zhou$^{1}$

\thanks{Manuscript received: September 3, 2024; Revised: December 11, 2024; Accepted: January 15, 2025.}
\thanks{This paper was recommended for publication by Editor Markus Vincze upon evaluation of the Associate Editor and Reviewers' comments.}
% \thanks{The project was supported by the NSF Grant RI-2339769 and the Sony Focused Research Award.}
\thanks{$^{1}$ University of California, Los Angeles {(\tt \footnotesize bolei@cs.ucla.edu} \newline {\tt \footnotesize matthewleng@cs.ucla.edu)}}
\thanks{$^{2}$ University of Southern California {(\tt \footnotesize yangjiaw@usc.edu)}}
\thanks{$^{3}$ Aurora Innovation {(\tt \footnotesize zren@aurora.tech)}}
\thanks{$^{\dagger}$ All work done while at the University of California, Los Angeles}
\thanks{Digital Object Identifier (DOI): see top of this page.}
}

\title{BEVCon: Advancing Bird's Eye View Perception with Contrastive Learning}


\maketitle

\begin{abstract}
We present BEVCon, a simple yet effective contrastive learning framework designed to improve Bird’s Eye View (BEV) perception in autonomous driving. BEV perception offers a top-down-view representation of the surrounding environment, making it crucial for 3D object detection, segmentation, and trajectory prediction tasks. While prior work has primarily focused on enhancing BEV encoders and task-specific heads, we address the underexplored potential of representation learning in BEV models. BEVCon introduces two contrastive learning modules: an instance feature contrast module for refining BEV features and a perspective view contrast module that enhances the image backbone. The dense contrastive learning designed on top of detection losses leads to improved feature representations across both the BEV encoder and the backbone. Extensive experiments on the nuScenes dataset demonstrate that BEVCon achieves consistent performance gains, achieving up to +2.4\% mAP improvement over state-of-the-art baselines. Our results highlight the critical role of representation learning in BEV perception and offer a complementary avenue to conventional task-specific optimizations. Code and models are available at \url{https://github.com/matthew-leng/BEVCon}. 
\end{abstract}

\begin{IEEEkeywords}
Object Detection, Segmentation, and Categorization; Deep Learning for Visual Perception; Representation Learning.
\end{IEEEkeywords}

\vspace{-1em}
\section{Introduction}
\label{sec:intro}
\IEEEPARstart{I}{n} recent years, Bird's Eye View (BEV) perception has emerged as a crucial component in autonomous driving and robotic systems \cite{pan2020cross,10.1007/978-3-030-58568-6_12,yang2022bevformer,blumenkamp2024covis}. Its ability to aggregate multi-view data and transform the surrounding environment in a unified top-down-view representation makes it highly effective and versatile for tasks like object detection, segmentation, trajectory prediction, and planning \cite{huang2021bevdet, liu2023bevfusion}. A typical BEV perception model architecture comprises an image backbone, a BEV encoder, and task-specific heads \cite{harley2023simple}. While many prior efforts have focused on optimizing the design of BEV encoders and task heads to improve performance \cite{Li_Ge_Yu_Yang_Wang_Shi_Sun_Li_2023,liu2022petr}, much less attention has been paid to enhancing BEV perception from a representation learning perspective. We argue that learned representations are central to a model's performance, and improving them can lead to uniform gains across various BEV architectures, offering broader benefits complementary to task-specific designs.

This work tackles this gap by integrating contrastive learning into BEV perception. Contrastive learning, which has shown remarkable success in the natural image domain through self-supervised learning approaches~\cite{chen2020simple,he2020momentum}, has recently demonstrated substantial potential in supervised settings as well~\cite{khosla2020supervised, balasubramanian2022contrastive, yu2023icpc}. By leveraging contrastive learning, we seek to enhance the underlying feature representations in BEV perception models, thereby improving their performance on downstream tasks like object detection. This introduces a new perspective of harnessing additional learning objectives beyond conventional detection losses without extra annotations, enriching the learning process.

However, incorporating contrastive learning into BEV representation learning presents many challenges. To investigate whether a naive application of contrastive learning could improve BEV performance, we conduct preliminary experiments by both pre-training (without detection losses) and co-training (with detection losses) the image backbones on various datasets and report results in \cref{table:contrast}.  Surprisingly, we observe that conventional contrastive learning methods, which focus on object-centric, image-level contrasts, fail to improve BEV detection performance. On average, pre-training methods result in 1.3\% mAP decrease, while co-training only yields 1.4\% mAP increase. These demonstrate that neither pre-training the image backbone with contrastive losses nor combining detection losses with image-level contrastive losses results in noticeable gains. We attribute the lack of improvement to two main factors: (1) Sample diversity: Driving \cite{nuscenes, Sun_2020_CVPR, NEURIPS_DATASETS_AND_BENCHMARKS2021_4734ba6f} and robotics datasets \cite{walke2023bridgedata, fang2024rh20t}  tend to exhibit much lower sample diversity compared to large-scale general-purpose datasets like ImageNet~\cite{imagenet_cvpr09}. These datasets typically contain a limited variety of object classes and backgrounds, making it challenging to learn sufficiently discriminative features through contrastive learning. (2) Level of contrasting: Image-level contrast methods focus on entire images rather than specific objects or details. This approach is effective for object-centric datasets like ImageNet, where the object of interest occupies most of the image. However, it becomes less effective in BEV perception tasks, where wide field-of-view images contain multiple objects and intricate details across the scene. The lack of focus on object-level contrasts limits the ability to capture fine-grained features.

To overcome these challenges, we introduce BEVCon, a dense contrastive learning framework to enhance BEV representation learning that can be seamlessly integrated into existing approaches. BEVCon consists of two contrast modules: (1) an instance feature contrast module, which enhances BEV feature extraction by performing dense contrastive learning on BEV features lifted from multi-view image data, leveraging annotations to improve localization and discriminativeness for detection tasks; and (2) a perspective view contrast module, which guides image backbone by focusing on region-specific features associated with different objects in the scene, helping the backbone capture more fine-grained, instance-level details. Together, these modules provide an effective strategy for learning better BEV representations.

We evaluate the effectiveness of our BEVCon through extensive experiments, instantiating it with three representative families of state-of-the-art BEV perception methods: depth-unprojection-based methods such as BEVDet~\cite{huang2021bevdet}, dense query-based methods like BEVformer~\cite{10.1007/978-3-031-20077-9_1}, and sparse query-based methods such as Sparse4D~\cite{lin2022sparse4d}. Our results show that the framework significantly enhances the learned representations of both the BEV encoder and the image backbone for BEV detection. Specifically, on the competitive nuScenes detection benchmark 
\cite{nuscenes}, it achieves a 2.4 mAP gain over the \emph{BEVFormer-tiny} model and a 1.3 mAP gain over the \emph{Sparse4D$_{T=1}$} model. Moreover, it consistently generalizes these improvements across multiple BEV methods.

Our contributions can be summarized as follows: 
\begin{itemize}
    \item We introduce contrastive learning into BEV detection models with two specialized contrastive learning modules, providing additional training objectives that enhance BEV representation learning.
    \item We design a unified framework that integrates these modules with detection tasks, aligning their optimization to enable joint training and improving detection performance.
    \item We conduct extensive experiments to validate the applicability and generalizability of our BEVCon across various BEV approaches, achieving consistent improvements.
\end{itemize}
Our study highlights that well-learned BEV representations lead to universal improvements across multiple BEV architectures and should receive more attention. This offers a complementary path to task-specific design optimizations.

\vspace{-1em}
\section{Related Work}

\paragraph{BEV Perception} Visual perception is one of the most important components of autonomous vehicles and robotics, including tasks like 3D object detection, semantic segmentation, and object tracking. Traditional perception algorithms perform 3D object detection independently on multiple cameras and then fuse those outputs in the object tracking component \cite{10.1007/978-3-030-58568-6_12}. Recently, Bird's-Eye View (BEV) perception has gained rising attention since it constructs a unified top-down-view feature map from perspective-view images for downstream perception tasks.
Jonah \textit{et al.} \cite{10.1007/978-3-030-58568-6_12} proposes a novel algorithm that converts image data from different cameras into a BEV scene representation, by first generating 3D features from 2D image features using a latent depth distribution, then flattening those 3D features to BEV feature maps. Huang \textit{et al.} \cite{huang2021bevdet} develops a similar framework of BEV feature map generation to 3D object detection tasks in autonomous driving and achieves significant performance gain. Different from the above two bottom-up approaches, Wang \textit{et al.} \cite{wang2022detr3d} and Li \textit{et al.} \cite{10.1007/978-3-031-20077-9_1} use top-down approaches to build BEV representations by leveraging transformers and matching 3D object queries with 2D image features to make bounding box predictions.
Liu \textit{et al.} \cite{liu2023bevfusion} fuses the learned BEV representations from image features with LiDAR BEV features into a multi-modality representation.

Most of the recent works in BEV perception contain one or more of the following items: temporal information aggregation from historical frames \cite{can2022understanding, 10.1007/978-3-031-20077-9_1, liu2023petrv2, wang2023exploring, 10.1609/aaai.v37i1.25185}, depth supervision \cite{Li_Ge_Yu_Yang_Wang_Shi_Sun_Li_2023, huang2022tigbev, zhang2023sabev, Li2023FBBEV}, or powerful pre-trained image backbones \cite{liu2023bevfusion, yang2022bevformer, Park2022TimeWT} (\textit{e.g.} using Swin~\cite{liu2021swin} or InternImage~\cite{wang2023internimage}). However, it remains much less explored if we can fully exploit the information in the training data from representation learning. In many real-world scenarios, it is difficult to gather external data, while hardware constraints could limit the incorporation of heavy pre-trained image backbones. Hence, it is important to investigate how to fully utilize the information in the
existing image \& BEV features and the corresponding annotations. Our work incorporates contrastive learning as a key gradient to utilize the existing image training data in BEV representation learning and yield better performance without adding extra supervision signals or external data. 

\paragraph{Contrastive Learning} Contrastive learning is a popular self-supervised learning approach. 
In image recognition, contrastive learning has effectively improved the features from unlabelled image data \cite{chen2020simple, he2020momentum, grill2020bootstrap, zbontar2021barlow}. Although several works have shown that using contrastive learning in supervised or semi-supervised context improves the performance of certain tasks \cite{khosla2020supervised, alonso2021semi, zhong2021pixel}, incorporating contrastive learning to BEV representation learning remains largely under-explored. Xia \textit{et al.} \cite{xia2023coin} attempts to use instance-level contrastive learning on BEV features generated from point clouds to help semi-supervised learning in 3D object detection. However, their method mainly focuses on addressing poor performance under sparse label settings, which only works on LiDAR and does not involve any image input. Our work proposes a contrastive learning framework that addresses the two unique challenges we discovered on image-based BEV representation learning through two newly introduced modules, with improved performance of 3D detection for autonomous vehicles and robotics by a significant margin across various BEV perception backbones.
\label{sec:related_work}

\vspace{-1em}
\section{Method}
We introduce BEV Contrastive Learning (BEVCon), a framework that harnesses contrastive learning to learn high-quality BEV features for autonomous vehicles. 
This section describes the architecture of our proposed framework, as well as our algorithm details. (\S \ref{sec:preliminary}) gives a brief introduction to the preliminaries of BEV representations and the limitations of existing works. (\S \ref{sec:bevcon}) presents an overview of our design. (\S \ref{subsec:ins_con}) and (\S \ref{subsec:pers_con}) details how our algorithm addresses the limitations of the BEV encoder and the image backbone. Lastly, we describe the joint optimization process in (\S \ref{sec:optim}).

\begin{figure}[t]
    \centering
    \includegraphics[width=0.45\textwidth]{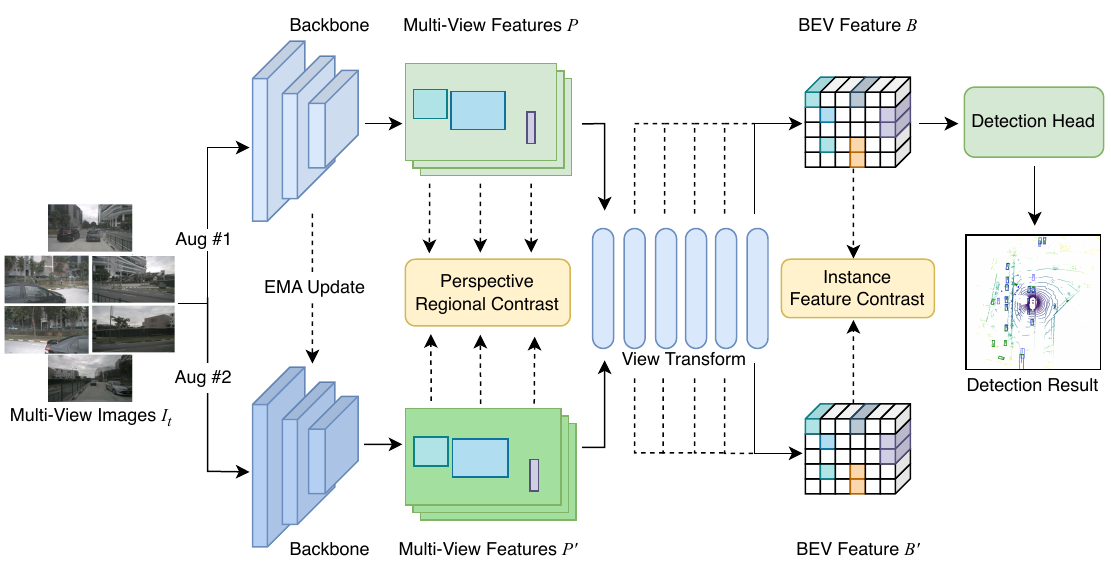}
   \caption{The overall framework of BEVCon, which consists of an instance feature contrast module and a perspective regional contrast module. It combines the losses from these two modules with the primary detection task. 
   }
    \vspace{-1.5em}
   \label{fig:arch}
\end{figure}

\vspace{-1em}
\subsection{Preliminaries}
\label{sec:preliminary}

\paragraph{BEV representations} 
Bird's Eye View (BEV) representation is a world representation that offers a top-down-view feature of the surrounding environment for an agent, such as an autonomous vehicle and other mobile robots. Denoted as $B \in \mathbb{R}^{H \times W \times C}$, a BEV representation can be typically formulated as a discretized grid structure feature with a $H \times W$ shape. BEV can be easily generated using data from the LiDAR and radar sensors. An image-based BEV representation can be obtained through a view transform module that extracts BEV features from image features, which is introduced as follows.

\paragraph{View transform.}
View transform is the process that converts multi-camera features from 2D to the BEV space. Let $P = \{P_j \, | \, j=1, ..., N_\text{level}\}$ be the image features extracted from different levels of an image backbone (e.g., different stages of a ResNet backbone), where $P_j = \{P_j^k \, | \, k=1,...,N_\text{view}\}$ denotes $N_\text{view}$ multi-camera features at level $j$. The view transform aims at learning a mapping $\mathcal{M}$ from image features $P$ to BEV feature $B$, $i.e.$, $B = \mathcal{M}(P)$. The actual form of this mapping varies across different methods. For example, forward-projection-based methods \cite{10.1007/978-3-030-58568-6_12} utilize depth prediction to unproject multi-view feature to BEV space, while backward-projection-based methods \cite{10.1007/978-3-031-20077-9_1} use object queries and project their center back to image space. In addition, positional encoding \cite{liu2022petr, liu2023petrv2} is used to encode image features and map the feature through an attention mechanism. 
However, the learning process of modern view transformation modules could be challenging. It usually requires a longer training time to converge \cite{yang2022bevformer}, and sometimes it might lose certain fine-grained information useful for downstream tasks \cite{Hu_2023}.

\begin{figure}[t]
    \vspace{-1em}
    \centering
    \includegraphics[width=0.48\textwidth]{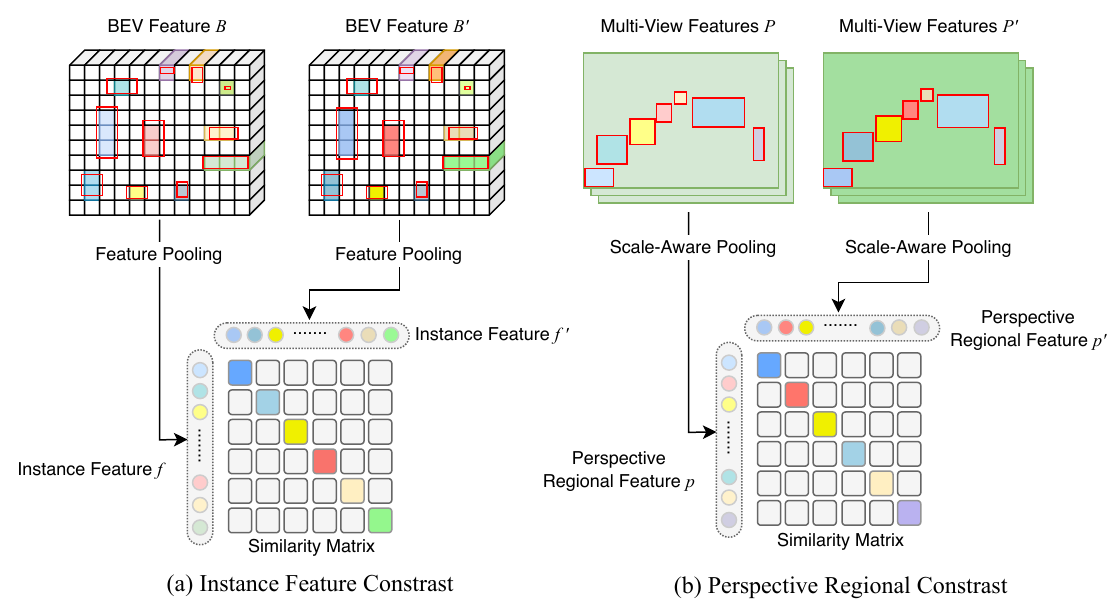}
    \caption{Illustration of two contrastive learning modules. Red bounding boxes in (a) denote the ground truth annotations in BEV space. 2D bounding boxes in (b) correspond to annotations in perspective view.}
    \label{fig:contrast_module}
     \vspace{-1.5em}
\end{figure}

\subsection{BEVCon}
\label{sec:bevcon}

Our method aims to improve the representation learning for BEV perception tasks.
As depicted in \cref{fig:arch}, BEVCon contains two components: an Instance Feature Contrast module that refines BEV features, and a Perspective Regional Contrast module that offers direct supervision for training image backbone. 

Given a set of multi-camera images $I_{t} = \{I_{t}^k \, | \, k = 1, ... N_\text{view} \}$ at timestamp $t$, we first transform them by applying data augmentation to form a pair of contrastive samples $(I, I')_{t}$. 
Subsequently, their respective features $(P, P')_{t}$ are extracted via a pair of Exponentially Moving Average (EMA) updated image backbones, which are then transformed into a pair of BEV representations $(B, B')_{t}$ utilizing a shared view transform module.
Apart from learning the BEV representation to only minimize detection loss, we augment the primary supervision signal with two contrastive learning modules to facilitate the learning of the image backbone and the view transform module simultaneously. We introduce each contrast learning module below.

\vspace{-1em}
\subsection{Instance Feature Contrast}
\label{subsec:ins_con}

Lifting image features to the BEV space is a challenging task. Taking BEVformer~\cite{10.1007/978-3-031-20077-9_1} as an example, it employs multiple transformer layers with learnable queries~\cite{carion2020endtoend} to accomplish this process, alongside a deformable attention module to alleviate the computational overhead. However, the gradient flow from the detection loss to this view transformation process has to go through a stack of transformer layers and recursive learnable queries, which could result in distortion for the supervision signal. Moreover, the absence of depth information leads to uniform features along each visual ray, producing ambiguous distance prediction and ray-shaped features \cite{Li2023FBBEV}, which affect the quality of BEV representation and detection performance.

To address these issues, we propose an instance feature contrast module that provides dense supervision for discrimination and alleviates the localization noise, as shown in \cref{fig:contrast_module}.
Specifically, given a set of two BEV representations ${B}$ and ${B'}$ with shape $h\times w$ regarding two augmented views of the same frame, and a set of $n$ annotations ${A}$ corresponding to that frame, we can extract feature sets for instances $f = \{ \mathcal{FP} \left( {B}, {A}_i \right) \, | \, i=1, ..., n \}$ through the feature pooling operation $\mathcal{FP}$. This process is repeated for $f' = \{ \mathcal{FP} \left( {B'}, {A}_i \right) \, | \, i=1, ..., n \}$. Our objective is to maximize the similarity of positive feature pair $\{\mathrm{similarity}\left( f_i, f'_i \right) \, | \, i=1, ..., n\}$ from the same instance while minimizing the similarity between negative pairs $\{ \mathrm{similarity}\left( f_i, f'_j \right) \, | \, i=1, ..., n, j=1, ..., n, i\neq j\}$.

Motivated by SimCLR \cite{chen2020simple}, we treat each instance feature from either augmented view as a sample without explicitly negative pair sampling, resulting in $2n$ sample points. For any positive pair defined above, we use the remaining $(n-1)$ as negative samples. The loss function, using cosine similarity for distance measurement, for a positive pair $(i, i)$ is defined as:
\begin{equation}
    \ell_{i} = -\log \frac{\exp(f_i \cdot f'_i/\tau)}
    {\sum_{j\neq i} \exp(f_i \cdot f'_j/\tau)},
    \label{eq:ins_loss}
\end{equation}
where $\tau$ is a temperature hyper-parameter. Note that we always normalize the feature vectors so that the cosine similarity is the dot-product of two feature vectors. We define the instance feature loss $\mathcal{L}_{in}$ as the average of the above loss over a batch sample with total $N$ instances.
% \begin{equation}
%     \mathcal{L}_{in} = \frac{1}{N} \sum_{k=1}^N \ell_{k}.
%     \label{eq:ins_loss_total}
% \end{equation}

For a dense query-based view transform module comprising of $N_\text{layer}$ transformer layers like BEVFormer \cite{10.1007/978-3-031-20077-9_1}, the intermediate output of the $l^{th}$ layer $(B, B')^l_t$ obtained from a set of augmented views $(I, I')_t$ can also form a contrastive pair for the instance feature contrast module. To modulate the impact of instance contrasting across various layers, we adopt an exponential scaling approach for the corresponding instance contrasting loss $\mathcal{L}^l_{in}$ regarding the layer $l$. The formulation of the instance feature contrast module's loss is expressed as follows:
\begin{equation}
    \mathcal{L}_{in} = \sum_{l} \frac{1}{\epsilon^{N_\text{layer} - l}} \mathcal{L}^{l}_{in},
\end{equation}
where $\epsilon$ is the multi-layer loss scale hyper-parameter.

One critical issue is the mismatch between annotation resolution and BEV grid size. Limited by the model scale and computation cost, a BEV representation typically has a limited discretized grid size (e.g. from 2 meters to 0.5 meters). Naively treating grid features as sample features will bring feature misalignment when the BEV grid size is large, resulting in highly imprecise local features. To address this issue, we use region-of-interest (RoI) Align \cite{he2017mask} to extract object features from low-resolution BEV feature maps.

\begin{table*}[t]
    \centering
    \caption{Generalization of BEVCon to various BEV methods on nuScenes $val$ set. $^\ddagger$ is initialized from backbone supervised pre-trained on ImageNet. $^\dagger$ is initialized from a FCOS3D backbone. $^*$ with CBGS strategy. $^\mathsection$ with a single frame.}
    \resizebox{1.5\columnwidth}{!}{
    \begin{tabular}{l|c|c|cc|ccccc}
        \toprule
        Method & Backbone  & Image Size & NDS $\uparrow$ & mAP $\uparrow$ & mATE $\downarrow$ & mASE $\downarrow$ & mAOE $\downarrow$ & mAVE $\downarrow$ & mAAE $\downarrow$ \\ 
        \midrule
        BEVDet$^\ddagger$  & \multirow{2}{*}{ResNet-50} & \multirow{2}{*}{800 $\times$ 1300} & 0.350 & 0.283 & 0.773 & 0.288 & 0.698 & 0.864 & 0.291 \\
        BEVDet+BEVCon$^\ddagger$  &  &  & \textbf{0.360} & \textbf{0.286} & \textbf{0.754} & \textbf{0.283} & \textbf{0.676} & \textbf{0.852} & \textbf{0.269} \\
        \midrule
        BEVDet4D$^{* \ddagger}$  & \multirow{2}{*}{ResNet-50} & \multirow{2}{*}{800 $\times$ 1300} & 0.447 & 0.314 & \textbf{0.691} & 0.282 & 0.549 & \textbf{0.381} & \textbf{0.196} \\
        BEVDet4D+BEVCon$^{* \ddagger}$  &  &  & \textbf{0.451} & \textbf{0.320} & 0.694 & \textbf{0.276} & \textbf{0.517} & 0.393 & 0.203 \\
        \midrule
        BEVFormer$^\ddagger$  & \multirow{2}{*}{ResNet-50} & \multirow{2}{*}{450 $\times$ 800} & 0.354 & 0.252 & 0.900 & 0.294 & 0.655 & 0.657 & \textbf{0.216} \\
        BEVFormer+BEVCon$^\ddagger$  &  &  & \textbf{0.375} & \textbf{0.276} & \textbf{0.872} & \textbf{0.284} & \textbf{0.634} & \textbf{0.621} & \textbf{0.216} \\
        \midrule
        Sparse4D$^{\dagger \mathsection}$   & \multirow{2}{*}{ResNet-101-DCN} & \multirow{2}{*}{640 $\times$ 1600} & 0.451 & 0.382 & 0.710 & 0.279 & 0.411 & 0.806 & \textbf{0.196} \\
        Sparse4D+BEVCon$^{\dagger \mathsection}$  &  &  & \textbf{0.460} & \textbf{0.395} & \textbf{0.709} & \textbf{0.276} & \textbf{0.397} & \textbf{0.787} & 0.208 \\
        \bottomrule
    \end{tabular}
    }
    \label{table:generalization}
    \vspace{-1em}
\end{table*}
\begin{table*}[t]
    % \vspace{-1em}
    \centering
    \caption{3D detection results of various methods on nuScenes $val$ set. $^\dagger$ is initialized from a FCOS3D backbone.}
    \resizebox{1.5\columnwidth}{!}{
    \begin{tabular}{l|c|c|cc|ccccc}
        \toprule
        Method & Backbone  & Image Size & NDS $\uparrow$ & mAP $\uparrow$ & mATE $\downarrow$ & mASE $\downarrow$ & mAOE $\downarrow$ & mAVE $\downarrow$ & mAAE $\downarrow$ \\ 
        \midrule
        BEVDet & ResNet-101 & 384 $\times$ 1056 & 0.396 & 0.330 & 0.702 & 0.272 & 0.534 & 0.932 & 0.251 \\
        PolarFormer$^\dagger$ & ResNet-101-DCN & 800 $\times$ 1333 & 0.458 & 0.396 & 0.700 & 0.269 & 0.375 & 0.839 & 0.245 \\
        DETR3D$^\dagger$ & ResNet-101-DCN & 900 $\times$ 1600 & 0.434 & 0.349 & 0.716 & \textbf{0.268} & 0.379 & 0.842 & 0.200 \\
        Sparse4D$^{\dagger}$   & ResNet-101-DCN & 640 $\times$ 1600 & 0.451 & 0.382 & 0.710 & 0.279 & 0.411 & 0.806 & 0.196 \\
        BEVFormer$^\dagger$ & ResNet-101-DCN & 900 $\times$ 1600 & 0.517 & 0.415 & \textbf{0.672} & 0.274 & 0.369 & 0.397 & 0.198 \\
        BEVFormer+BEVCon$^\dagger$ & ResNet-101-DCN  & 900 $\times$ 1600 & $\textbf{0.528}$ & $\textbf{0.424}$ & 0.674 & 0.274 & $\textbf{0.357}$ & $\textbf{0.354}$ & $\textbf{0.183}$ \\
        \bottomrule
    \end{tabular}
    }
     \vspace{-1.5em}
    \label{table:val}
\end{table*}

\vspace{-1em}
\subsection{Perspective Regional Contrast}
\label{subsec:pers_con}
As discussed before, image-based BEV algorithms rely on learned view transform modules to map the features from 2D images to the BEV space. If not well learned, those view transform modules can distort the gradient flow from both the detection head and the instance feature contrast module, leading to sparse supervision for the image backbone. We will tackle this issue in this section.

To augment the supervision signals and the gradient flows for the image backbone, we develop a perspective regional contrast module that performs dense contrasting on the RoI image features, as illustrated in \cref{fig:contrast_module}.
Given multi-level features from image backbone $P = \{P_j^k \, | \, k=1,...,N_\text{view}, j=1, ..., N_\text{level}\}$ corresponding to a total of $N_\text{view}$ camera views and $N_\text{level}$ output levels, alongside 2D annotations $A^{2D}$ for the frame. Similar to the instance feature contrast module, we obtain features of level $j$ from the annotation region as $p_j = \{\mathcal{FP}^{2D} \left( P_j, A^{2D}_i \right) \, | \, i=1,...,n \}$. These features, when contrasted with features $p'_j$ from an augmented view, provide direct supervision for image features.

Another issue is that the bounding box annotations in driving scenes usually overlap. For example, the bounding box of a vehicle closer to the ego car will likely cover other objects behind that vehicle. Naively pooling the features within a bounding box will inevitably and undesirably fuse features of different objects together. To address this problem, we propose a scale-aware pooling mechanism parameterized by a scale parameter $\gamma$. It downscales the bounding box annotations by the pre-defined scale parameter so that only the central features will be pooled. This approach leads to more precise object features.

Lastly, to ensure the different stages of the image backbone all receive proper dense supervision, we propose to perform multi-level dense contrasting. Specifically, we apply the perspective regional contrastive loss to different stages and obtain the perspective contrast loss $\mathcal{L}_{pers}$ by averaging them.
% \begin{equation}
%     \mathcal{L}_{pers} = \frac{1}{N_\text{level}} \sum_{j=1}^{N_\text{level}} \mathcal{L}_{pers}^{j}.
% \end{equation}

\vspace{-1em}
\subsection{Joint Optimization}
\label{sec:optim}

We design two losses to improve the view transform module and the image backbone, two critical components in BEV perception systems, alongside the original detection supervision losses. 

By combining the loss $\mathcal{L}_{in}$ which serves as supervision for the view transform module and image backbone, and the loss $\mathcal{L}_{pers}$ which provides direct supervise to image backbone, the joint training loss that enables the synergy is defined as
\begin{equation}
    \mathcal{L} = \lambda_{in} \mathcal{L}_{in} + \lambda_{pers} \mathcal{L}_{pers} + \mathcal{L}_{det},
\end{equation}
where $\mathcal{L}_{det}$ contains the losses of detection task.

\vspace{-1em}
\section{Experiments}

\subsection{Dataset and Metrics}

We evaluate our method on the widely benchmarked nuScenes dataset \cite{nuscenes}. It contains a total of 1000 driving sequences, each of 20s. Each sequence consists of multi-modal information and 2$Hz$ annotations, where each sample includes RGB images from 6 cameras which offer a full surrounding view field. The sequences are split into 700, 150, and 150 for training, validation, and testing, respectively. For the 3D detection task, the benchmark contains roughly 1.4M 3D object bounding boxes across 10 classes of objects. For evaluation, we use the provided evaluation metrics, including mean Average Precision (mAP), nuScenes Detection Score (NDS), and 5 true positive metrics mean Average Translation Error (mATE), mean Average Scale Error (mASE), mean Average Orientation Error (mAOE), mean Average Velocity Error (mAVE), and mean Average Attribute Error (mAAE).

\vspace{-1.9em}
\subsection{Implementation Details}

We evaluate our contrastive learning framework on three families of image-based BEV methods with different view transformation strategies and training settings:

\begin{enumerate}
    \item Depth-based methods including BEVDet and BEVDet4D \cite{huang2021bevdet, huang2022bevdet4d}. We perform feature pooling on the BEV representation attained from projections and the perspective view feature for contrastive learning while maintaining other settings identical to the papers. Models are trained with AdamW optimizer and a learning rate of $2 \times 10^{-4}$ for 20 epochs under 64 batch size.
    \item Dense query-based method such as BEVFormer \cite{10.1007/978-3-031-20077-9_1}. We use two models of the BEV detector BEVFormer \cite{10.1007/978-3-031-20077-9_1}, \emph{BEVFormer-tiny} and \emph{BEVFormer-base}, following the original settings. One with ResNet-50 \cite{he2016deep} as the image backbone, 3 BEV encoder layers for view transform module which produces 50 $\times$ 50 resolution BEV feature. The other one uses ResNet-101 \cite{he2016deep}, 6 BEV encoder layers which outputs 200 $\times$ 200 BEV feature. We utilize BEV features from multiple encoder layers and the perspective view feature for contrastive learning. All models are trained with 24 epochs, a learning rate of $2 \times 10^{-4}$, a batch size of 8, and gradient clipping.
    \item Sparse query-based method such as Sparse4D \cite{lin2022sparse4d}. We adapt our framework to Sparse4D by using instance features extracted by sparse query embeddings for contrastive learning. The AdamW optimizer is used with cosine annealing and gradient clipping, and the initial learning rates of 2e-5 and 2e-4 for the backbone and other parameters, respectively. All experiments are trained with 24 epochs and a batch size of 8.
\end{enumerate}

For contrastive learning augmentations, we adopt widely used random resize, rotation, flipping, cropping, and distortion \cite{chen2020simple} for the input image and additional random rotation, scaling, and flipping regarding the BEV representations to further extend the spatial augmentation. The whole augmentation process ensures that each instance feature pair regarding the same instance is located in the same position in BEV space, which enables contrastive learning. We set loss factor $\lambda_{in}$ and $\lambda_{pers}$ to 1, the center scale factor $\gamma$ to 0.6, and multi-layer loss scale $\epsilon$ to 0.5. The contrastive temperature $\tau$ is set respectively to 0.2 and 0.05 for the two image backbones. Contrastive loss is combined with the baseline's loss for joint optimization.

\vspace{-1em}
\subsection{Results}

\begin{table*}[t!]
    \centering
    \caption{3D detection results of different designed modules applied to BEVFormer on the nuScenes $val$ set.}
    \vspace{-0.5em}
    \resizebox{1.6\columnwidth}{!}{
    \begin{tabular}{c|c|cc|ccccc}
        \toprule
        Setting & Backbone  & NDS $\uparrow$ & mAP $\uparrow$ & mATE $\downarrow$ & mASE $\downarrow$ & mAOE $\downarrow$ & mAVE $\downarrow$ & mAAE $\downarrow$   \\ 
        \midrule 
        No Contrast & \multirow{4}{*}{ResNet-50} &  0.354 &  0.252 &  0.900 &  0.294 &  0.655 &  0.657 & $\textbf{0.216}$ \\ 
        Instance Contrast &  & $\textbf{0.375}$ (0.3748 $\pm$ 0.0002) & 0.272 (0.2700 $\pm$ 0.0009) &  0.888 &  0.287 & $\textbf{0.587}$ & $\textbf{0.620}$ &  0.224 \\
        Perspective Contrast &  &  0.373 (0.3728 $\pm$ 0.0003) & 0.273 (0.2709 $\pm$ 0.0009) & $\textbf{0.865}$ &  0.294 &  0.601 &  0.643 &  0.228 \\
        Contrast Framework &  &  0.371 (0.3709 $\pm$ 0.0002) & $\textbf{0.274}$ (0.2735 $\pm$ 0.0002)  &  0.878 & $\textbf{0.286}$ &  0.640 &  0.642 &  0.218 \\
        \midrule 
        No Contrast & \multirow{4}{*}{ResNet-101-DCN} &  0.517 &  0.415 &  0.672 &  0.274 &  0.369 &  0.397 &  0.198 \\ 
        Instance Contrast &  & 0.523 (0.5213 $\pm$ 0.0011) & 0.420 (0.4197 $\pm$ 0.0003) &  0.675 & $\textbf{0.273}$ & $  0.358 $ &  0.375 & $\textbf{0.190}$ \\ 
        Perspective Contrast & &  0.522 (0.5215 $\pm$ 0.0016) & 0.418 (0.4168 $\pm$ 0.0009) & $\textbf{0.660}$ &  0.274 &  0.362 & $\textbf{0.363}$ &  0.201 \\ 
        Contrast Framework & & $\textbf{0.525}$ (0.5250 $\pm$ 0.0004) & $\textbf{0.422}$ (0.4212 $\pm$ 0.0005)  &  0.662 &  0.275 & \textbf{0.356} &  0.371 &  0.199 \\ 
        \bottomrule
    \end{tabular}
    }
    \vspace{-1em}
    \label{table:abl1}
\end{table*}
\let\ck\checkmark
\begin{table*}[t]
    % \vspace{-1em}
    \centering
    \caption{Ablation study on different components of BEVCon regarding different model sizes of BEVFormer on the nuScenes $val$ set. ``Ins'': instance contrast, ``Align'': RoI align pooling, ``MLC'': multi-layer features contrast, ``Pers'': perspective regional contrast, ``Scale'': scale-aware pooling. Baseline results are marked as {\color[HTML]{C0C0C0} gray}. }
\vspace{-1em}
\begin{adjustbox}{valign=c}
% \begin{subtable}[b]{0.45\textwidth}
    \begin{minipage}[b]{0.4\textwidth}
        \centering
        \caption*{(a) BEVFormer-tiny+BEVCon}
        \resizebox{\columnwidth}{!}{
            \begin{tabular}{ccc|cc|cc}
            \toprule
            Ins & Align & MLC & Pers & Scale & NDS $\uparrow$ & mAP $\uparrow$  \\ 
            \midrule
            & & &  &  & {\color[HTML]{C0C0C0} 0.354} & {\color[HTML]{C0C0C0} 0.252} \\ 
             \ck & & & & & 0.378 (0.3724 $\pm$ 0.0056) & 0.271 (0.2695 $\pm$ 0.0015)  \\ 
               \ck & \ck &  & & & 0.372 (0.3709 $\pm$ 0.0006) & 0.264 (0.2654 $\pm$ 0.0006) \\ 
               \ck & \ck & \ck & & & 0.375 (0.3748 $\pm$ 0.0002) & 0.272 (0.2700 $\pm$ 0.0009) \\ 
               & & & \ck & \ck & 0.373 (0.3728 $\pm$ 0.0003) & 0.273 (0.2709 $\pm$ 0.0009) \\ 
               \ck & \ck & \ck & \ck &  & 0.371 (0.3704 $\pm$ 0.0003) & 0.265 (0.2649 $\pm$ 0.0003)\\ 
               \ck & \ck & \ck & \ck & \ck  & 0.371 (0.3709 $\pm$ 0.0002) & 0.274 (0.2735 $\pm$ 0.0002) \\ 
            \bottomrule
            \end{tabular}
        }
        \label{tab:abl2_tiny}
    % \end{subtable}
    \end{minipage}
\end{adjustbox}
    \hspace{2em}
    % \hfill
\begin{adjustbox}{valign=c}
    % \begin{subtable}[b]{0.45\textwidth}
    \begin{minipage}[b]{0.4\textwidth}
        \centering
        \caption*{(b) BEVFormer-base+BEVCon}
        \resizebox{\columnwidth}{!}{
            \begin{tabular}{ccc|cc|cc}
            \toprule
            Ins & Align & MLC & Pers & Scale & NDS $\uparrow$ & mAP $\uparrow$  \\ 
            \midrule
             & & & & & {\color[HTML]{C0C0C0} 0.517} & {\color[HTML]{C0C0C0} 0.415} \\
             \ck & & & & & 0.519 (0.5184 $\pm$ 0.0003) & 0.417 (0.4162 $\pm$ 0.0004) \\
              \ck & \ck & \ck & & & 0.523 (0.5213 $\pm$ 0.0011) & 0.420 (0.4197 $\pm$ 0.0003)\\
               &  &  & \ck & \ck & 0.522 (0.5215 $\pm$ 0.0016) & 0.418 (0.4168 $\pm$ 0.0009)\\
              \ck & \ck & \ck  & \ck & \ck & 0.525 (0.5250 $\pm$ 0.0004) & 0.422 (0.4212 $\pm$ 0.0005)\\
            \bottomrule
            \end{tabular}
        } 
        \label{tab:abl2_base}
    % \end{subtable}
    \end{minipage}
\end{adjustbox}
\vspace{-1em}
\label{table:abl2_impl}
\end{table*}
\begin{table*}[t]
    % \vspace{-1em}
    \centering
    \caption{3D detection result on the nuScenes $val$ set based on \emph{BEVFormer-tiny}, our method is compared against: 1) contrastive pre-training using MoCo v2 for 100 epochs; 2) method combining detection loss with image-level contrastive loss. ``Pretrain Loss'' denotes the loss of MoCo v2 after pretraining.}
    \vspace{-0.5em}
    \resizebox{1.55\columnwidth}{!}{
    \begin{tabular}{l|c|c|cc|ccccc}
        \toprule
        Method & Training & Pretrain Loss & NDS $\uparrow$ & mAP $\uparrow$ & mATE $\downarrow$ & mASE $\downarrow$ & mAOE $\downarrow$ & mAVE $\downarrow$ & mAAE $\downarrow$ \\ 
        \midrule
        Pre-training on nuScenes & Pre-training & 6.40 & 0.338 & 0.224 & 0.928 & 0.296 & 0.659 & 0.642 & 0.212 \\
        Pre-training on ACO & Pre-training & 6.35 & 0.338 & 0.227 & 0.932 & 0.298 & 0.673 & 0.629 & 0.228 \\ 
        Image-level Contrast & Joint Training & - & 0.364 & 0.266 & 0.890 & 0.288 & 0.651 & 0.636 & 0.219 \\
        Pre-training on ImageNet & Pre-training & 6.60 & 0.371 & 0.266 & 0.883 & 0.292 & 0.643 & 0.588 & 0.211 \\
        BEVFormer+BEVCon (Ours) & Joint Training & - & 0.375 & 0.276 & 0.872 & 0.284 & 0.634 & 0.621 & 0.216 \\
        \bottomrule
    \end{tabular}
    }
    \label{table:contrast}
    \vspace{-1.5em}
\end{table*}

We first demonstrate the generalizability of our framework by applying it to the above three representative methods. \Cref{table:generalization} reports the results. We see that BEVCon consistently improves existing detectors without introducing any new data or labels. For BEV detectors using ResNet-50, it outperforms each of the original models, with up to 2.1\% NDS and 2.4\% mAP improvements on the baseline BEVFormer-tiny model. When incorporating it into the model with a larger backbone, our approach still brings 1.3\% mAP over Sparse4D. 

We conduct experiments to compare with existing state-of-the-art BEV detection methods including BEVDet \cite{huang2021bevdet}, BEVFormer \cite{10.1007/978-3-031-20077-9_1}, Sparse4D \cite{lin2022sparse4d},  PolarFormer \cite{10.1609/aaai.v37i1.25185}, and DETR3D \cite{wang2022detr3d}. To ensure a fair comparison, all benchmark experiments are running with the ResNet-101 backbone. \Cref{table:val} shows the 3D detection results of all the methods on the nuScenes $val$ set. By incorporating our proposed framework into BEVFormer using ResNet-101, it outperforms the original model with up to 1.1\% NDS and 0.9\% mAP improvements, surpassing these existing methods.

The experiments in \cref{table:generalization} and \cref{table:val} cover settings including different view transformation strategies, temporal fusion approaches, and training strategies. Our BEVCon demonstrates strong generalizability across different BEV methods with consistent improvements.

\vspace{-1em}
\subsection{Ablation Study}

The statistical performance results of BEVCon presented below are obtained by conducting 5 training runs for the model with a ResNet-50 backbone and 3 training runs for the model with a ResNet-101-DCN backbone.

\subsubsection{Contrast Learning Modules}

To better understand our BEVCon, we ablate different design choices in \cref{table:abl1}, where Instance Contrast and Perspective Contrast correspond to the modules in \cref{subsec:ins_con} and \cref{subsec:pers_con}. The Contrast Framework stands for the combination of the above two.

Compared with the detector with no contrast learning module, incorporating either one of the contrast learning modules leads to a performance increase across different model sizes, and the combination of them could further improve the performance. On the model with ResNet-50 as the backbone, our method can yield a 1.7\% NDS and 2.2\% mAP increase over the detector without contrastive learning, with 2.2\% and 1.5\% decrease in mATE and mAOE which are closely related to prediction localization. When deploying on the model with a larger backbone and higher BEV resolution, it still outperforms the original detector with a margin of 0.7\% in NDS and mAP, with 1\% $\sim$ 3\% decrease across mATE, mAOE, and mAVE. These improvements reveal the effectiveness of our contrastive framework across different scales of models without any existing structure modifications, improving the BEV representation from both the BEV feature space and perspective view space. The decrease in mATE, mAOE, and mAVE illustrates how the contrastive framework facilitates the training of the image backbone and BEV encoder by learning information regarding localization, orientation, and velocity. 

When comparing the effects of our contrastive modules across models of different sizes, we observe that both contrastive modules provide strong supervision for smaller network configurations (e.g., BEVFormer-tiny), and the contrast framework further reduces the standard deviation of mAP. As the framework is deployed on models with an increased number of BEV encoder and task-specific head layers, the supervision provided by \textit{Perspective Contrast}, \textit{Instance Contrast}, and the \textit{Contrast Framework} progressively strengthens, leading to noticeable performance improvements.

\begin{figure}[htbp]
    \centering
    \includegraphics[width=0.48\textwidth]{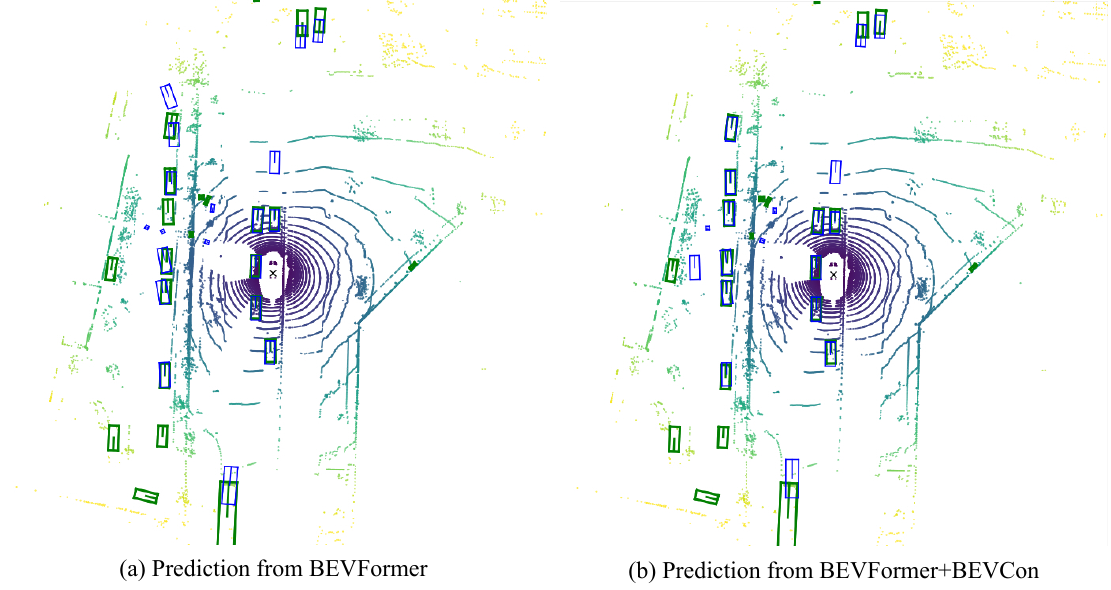}
    \caption{A qualitative comparison of the detection results between BEVFormer and BEVFormer+BEVCon. Green boxes correspond to GT annotations, blue boxes are the predicted bounding boxes. 
    }
    \label{fig:bev_pred}
     \vspace{-2em}
\end{figure}

\Cref{fig:bev_pred} presents the detection results from BEVFormer and our method. The incorporated contrastive learning improves the BEV representation from both the coarse and the fine-grained levels.
These improvements yield more precise predictions while mitigating the excessive predictions caused by ray-shaped artifacts.

\subsubsection{Sub-components of BEVCon}

\Cref{table:abl2_impl} contains an ablation study on the implementation choices of two modules separately and jointly. These include 1) RoI align feature pooling for Instance Feature Contrast to get more precise instance features. 2) Multi-layer instance contrast which performs regarding BEV features from different transformer layers in the view transform module. 3) Scale-aware pooling to mitigate the effect of overlapping RoI in image features. Comparing the results among models with only instance feature contrast, the RoI align and the multi-layer contrast provide more precise and strong supervision to the view transform module and further improve the BEV features. When performing perspective regional contrast, the results have shown scale-aware pooling as a crucial part, without which the model would encounter training instability and further affect the scale of the final improvement.

\subsubsection{Experiment on Contrastive Pre-training/Joint Training}
We further validate that traditional image-level contrastive learning is ineffective in boosting the performance on BEV-based object detection tasks.
For contrastive pre-training, we first pretrain the image backbone through unsupervised contrast learning using MoCo v2 \cite{chen2020improved} on ImageNet and two driving datasets, nuScenes \cite{nuscenes} and ACO \cite{zhang2022learning}, and then initialize the image backbone of our BEV detector with the pre-trained weights respectively. For joint training, we combine the contrastive loss from an image-level contrast module similar to MoCo v2~\cite{chen2020improved} with detection loss. Results in \cref{table:contrast} show that contrastive pre-training/joint training methods performed on driving scenes fall short of improving BEV representation and boosting detection performance. In particular, a smaller pretraining loss scale and a notable decrease in performance are observed from models initialized with driving dataset pre-trained weight. The smaller pretraining loss can be attributed to its lower sample diversity which makes the contrastive learning task trivial. The performance gain from pre-training on ImageNet falls short of our contrastive learning framework. One reason is that conventional unsupervised contrastive learning focuses on the whole
image rather than specific objects, causing a lack of specificity for learning object features.
These findings underscore the importance of contrast in BEV perception and elucidate the rationale behind our contrastive learning framework design.

\vspace{-1em}
\section{Conclusion}

This work presents a novel contrastive learning framework called BEVCon that can be seamlessly integrated into multiple representative BEV perception algorithms without access to external training data. By incorporating contrastive modules for both the BEV encoder and image backbone, the proposed method enhances representation learning by leveraging contrast signals alongside the primary detection task. Experimental results on the nuScenes benchmark demonstrate the effectiveness and generalizability of BEVCon, showcasing notable improvements over competitive baselines. This underscores the significance of representation learning for advancing BEV perception in autonomous vehicles and robotics.

\vspace{-1em}
\section*{Acknowledgment}

The project was supported by the NSF Grant RI-2339769 and the Sony Focused Research Award.

\vspace{-1em}
\bibliographystyle{IEEEtran}
\bibliography{IEEEabrv,IEEEexample}

\vfill
\end{document}